# Spine Vision – X-Ray Image based GUI Planning of Pedicle Screws Using Enhanced YOLOv5 for Vertebrae Segmentation

Yaswantha Rao P*, Gaurisankar S*, Durga R*, Aparna Purayath*, Vivek Maik*, Manojkumar Lakshmanan*, and Mohanasankar Sivaprakasam*†
Email: aparna_p@htic.iitm.ac.in
* Healthcare Technology Innovation Centre (HTIC), Indian Institute of Technology Madras, India
†Department of Electrical Engineering, Indian Institute of Technology Madras, India

***Abstract*—In this paper, we propose an innovative Graphical User Interface (GUI) aimed at improving preoperative planning and intra-operative guidance for precise spinal screw placement through vertebrae segmentation. The methodology encompasses both front-end and back-end computations. The front end comprises a GUI that allows surgeons to precisely adjust the placement of screws on X-Ray images, thereby improving the simulation of surgical screw insertion in the patient's spine. On the other hand, the back-end processing involves several steps, including acquiring spinal X-ray images, performing pre-processing techniques to reduce noise, and training a neural network model to achieve real-time segmentation of the vertebrae. The integration of vertebral segmentation in the GUI ensures precise screw placement, reducing complications like nerve injury and ultimately improving surgical outcomes. The Spine-Vision provides a comprehensive solution with innovative features like synchronous AP-LP planning, accurate screw positioning via vertebrae segmentation, effective screw visualization, and dynamic position adjustments. This X-ray image-based GUI workflow emerges as a valuable tool, enhancing precision and safety in spinal screw placement and planning procedures.**

## I. Introduction

The inception of Deep Learning (DL) has ushered in a paradigm shift within the healthcare industry [1] [2], particularly in the realm of Medical Imaging. The application of DL techniques has effectively automated previously manual and labor-intensive processes, offering substantial advancement. Within the domain of spine surgery, a complex and intricate field, the integration of DL and computer vision technologies assumes critical significance [3] [4]. The envisioned outcome is a reduction in the likelihood of human errors inherent in such complex procedures. Ma et al. [5] utilized the Faster-RCNN framework to identify cervical spinal cord injury and disc degeneration in spine MRIs. They achieved an impressive mean average precision (mAP) score of 88.6% and effectively demonstrated the ability to predict injuries. Although the results were encouraging, the utilization of Faster RCNN, a two-stage detector, resulted in rather lengthy inference times, rendering it unfeasible for real-time predictions [6] [7]. Lecron et.al [8] [9] introduced a method for segmenting vertebrae that uses parallel computing to compute diverse topologies. This reduces the expense of transferring data between memory units, making it suitable for handling extensive datasets and achieving enhanced accuracy. Mushtaq et.al [10] devised a system to localize and segment lumbar vertebrae. After localizing each vertebra from a sagittal spine x-ray, the specific area of interest is selected, and a corresponding mask is created. The image and the mask are inputted into the HED U-Net model to identify edges. Their performance on the dataset resulted in a mAP of 97.5%. However, this method does not address the coronal plane and synchronization between the two planes.

Researchers continuously enhance YOLOv5 for many medical imaging applications. Adji et al. [11] employed YOLOv5 along with the Weighted Box Fusion (WBF) method to achieve a mAP of 61% in detecting suspicious objects in thoracic X-rays. Although the model includes CLAHE pre-processing and WBF-based bounding box filtering, its accuracy is insufficient for real-world applications that require high precision. Zhang et al. [12] improved YOLOv5 by incorporating an attention mechanism and context information, resulting in a 3% increase in mAP at the intersection over union (IoU) threshold of 0.5 to 0.95, and a 5% increase in mAP at the IoU threshold of 0.5. Xia et al. [13] proposed the integration of Global Contextual Attention with Bi-directional Feature Pyramid Network (Bi-FPN) feature fusion, which successfully reduces interference from the background and improves the accuracy of feature extraction. Similarly, Xuan et al. [14] enhanced feature extraction in paddle paddle (PP-YOLOv2) by using Coord-Conv and Spatial Pyramid Pooling (SPP) modules, resulting in higher mAP scores when compared to YOLOv3 and YOLOv5. Guinebart et al. [15] employed UNET++ to do segmentation of vertebrae and used YOLOv5x for detection. They achieved significant mAP scores for both vertebrae and disk detection. Shan et al. [16] enhanced the performance of YOLO Tiny for detecting spinal fracture lesions in a different field. They achieved better accuracy by adding more convolutional layers, resulting in an excellent 85.63% mAP score. Additionally, they achieved a fast inference time of 0.0216s. The current exploration of image-based GUI planning for pedicle screw insertions is limited, as existing systems primarily rely on 3D CT or tools instead of 2D image-based planning. The proposed work introduces an innovative approach by integrating GUI features and enhancing the yolov5 algorithm for vertebrae segmentation and localization. The key innovations include the development of a synchronized GUI for planning in both top, anterior-posterior (AP) and side, lateral-posterior (LP) views, vertebrae segmentation at various levels using an improved yolov5 algorithm, and the creation of a customized YOLOv5 module with Adaptive Fusion (AF) and Co-ordinate Attention (CA) for precise vertebrae localization in 2D fluoroscopic images. This approach aims to improve accuracy and minimize errors in pedicle screw planning compared to existing methods that focus on 3D modalities.

## II. Methodology

### A. GUI Design and Working

The workflow begins with the acquisition of X-ray images of the spine from both AP and LP views. Subsequently, these images undergo pre-processing and augmentation to enhance quality and diversity. The annotated images form the training dataset for a deep learning model, which is then trained to segment vertebrae in X-ray images. The trained model is employed for inference and prediction on new X-ray images. In the surgical phase, the surgeon selects patient-specific AP and LP X-ray images. The GUI displays these images of the spine, allowing the surgeon to define the vertebra of interest for screw placement. Utilizing Z-correspondence, screws are added to the corresponding vertebral bounding box, determined from the segmentation module. The GUI facilitates a simulation of the surgical screws in the patient's spine, allowing the surgeon to adjust the placement as necessary. Following satisfactory screw placement, the GUI generates a surgical plan detailing screw size, type, and location. Fig(1), shows the GUI that employs vertebrae segmentation for planning and installing screws. [17] [18]

### B. Vertebrae Segmentation Using Improved YOLOv5

The initial step in the planning of pedicle screw placement in the spine involves uploading the AP & LP X-ray images, which provide a top view and side view of the spine. These images are then fed into the Spine-Vision model to segment the vertebrae on an in house dataset. The proposed Spine-Vision has two novel additions to the YOLOv5 architecture namely an AF and CA to enable it to learn complex features from Spine X-rays. Fig (2), shows the visual representation of Spine-Vision architecture, showcasing its backbone, neck, and head components for object detection. The incorporation of the AF module to the backbone is proposed, drawing inspiration from Adaptively Spatial Feature Fusion [19] and Spatial Pyramid Pooling [20]. The AF module aggregates local features from two different scales, followed by interpolation and feature fusion, like Spatial Pyramid Pooling (SPP), facilitating the detection of objects at various scales and capturing spatial information. This is crucial for X-ray images with varying contours in shape and size. Test results indicate that the AF module significantly aids in identifying smaller, intricate vertebrae in spine X-rays. While the increased spatial attention led to a slight decrease in predicted bounding box confidence scores for some test images, the incorporation of the CA module and spatial pyramid pooling fast (SPPF) [21] was introduced to mitigate this decrease. Fig (3), shows the implementation of the module. The CA module can capture input features along the vertical and horizontal directions. These feature maps are then multiplied with the input feature map, and this provides for better object localization. The CA module plays a crucial role in negating the confidence decrease caused by the AF module. The CA layer learns the directional and cross-channel information from the region of interest, which is important for improved object detection. Fig (4) shows the implementation of the CA module. SPPF is an optimized and faster version of Spatial Pyramid Pooling [22] that was originally used in YOLOv3 [20] . The SPPF module is used for multi-scale representation of the input feature maps. It incorporates feature fusion and pooling at various scales, which allows to capture of objects of various sizes. For each image, the model constructs a bounding box of vertebrae and records the top left and bottom right coordinates of the bounding box in a folder as a text file. [23]

### C. Pedicle Screw Placement and Planning

The pedicle screw is visualized as a geometric cylinder, a 3D object with dimensions in the X, Y, and Z axes. This cylinder will be represented in 2D space on the AP and LP images. The planning process starts with labeling the vertebrae of interest, followed by the back-end segmentation of the vertebra, initialization of the screw, screw visualization, selecting them, and adjusting their placements to achieve precise screw positioning. [24] The pedicle screw placement relies on the output of the enhanced YOLOv5 model and the screw is positioned within the bounding box of the labeled vertebra. After positioning, the surgeon can strategize the procedure by adjusting the screw either in the AP or LP image. When the screw is adjusted in either the AP or LP 2D space, the planning automatically updates on the corresponding 2D image and is synchronized, as illustrated in Fig (5). This synchronization is based on the concept that the 3D equivalent of the 2D object has a single axis that serves as a shared axis between the AP & LP images. Consequently, each movement performed in one on the 2D plot is mirrored in another plot through the shared common axis, by altering the value of the screw's common axis. The screw/cylinder is built on two points, the entry point and the target point where the entry point is the location on the spine where the surgeon makes an incision to insert the screw which is typically on the surface of the vertebra and serves as the starting point for the screw trajectory. The target point is the destination or endpoint within the vertebra where the surgeon aims to place the screw. In the 3D object space, the entry point and target point act as the centers of the two circular bases of the cylinder as shown in Figure 5.

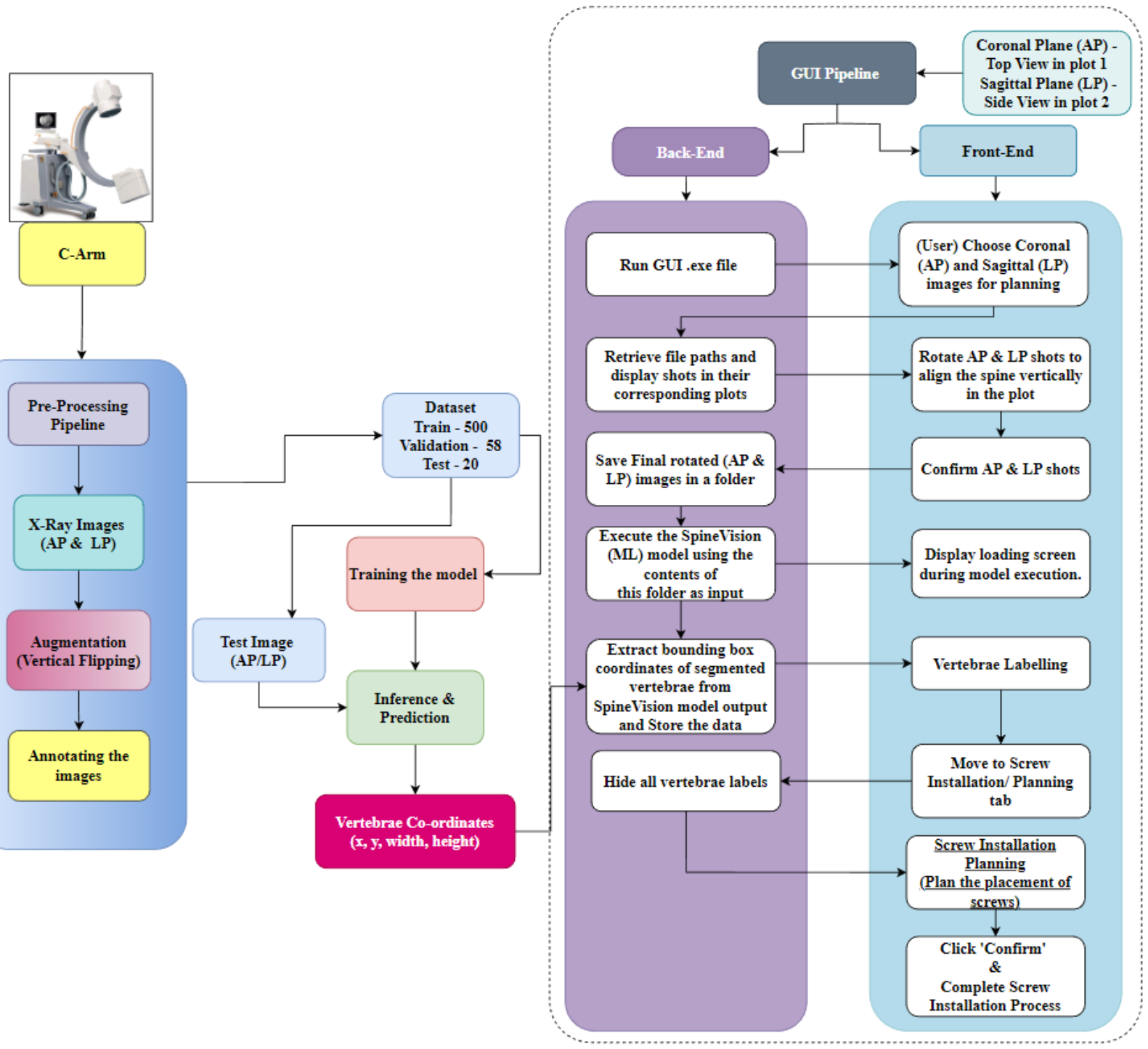

Fig. 1. Block diagram of proposed work: graphical user interface (GUI) for planning and installing screws in vertebrae using 3D segmentation

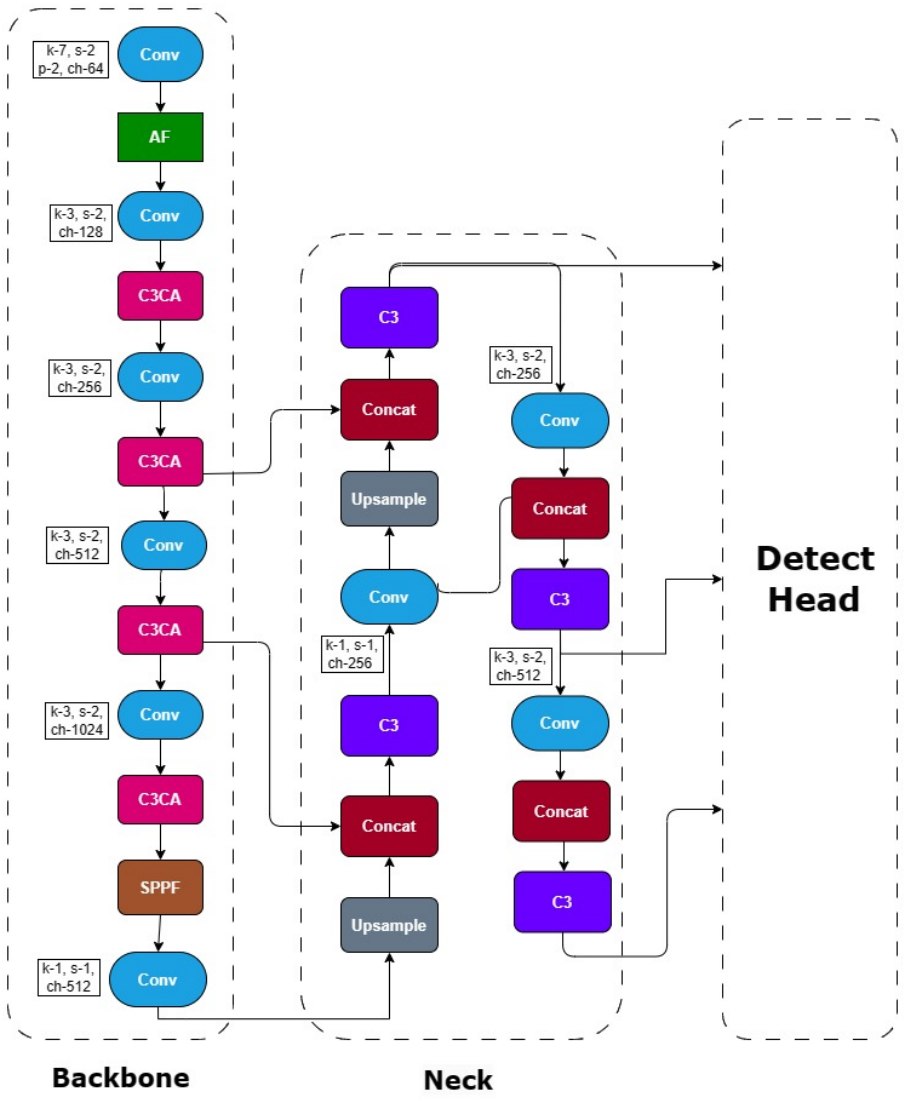

Fig. 2. Spine-Vision Model Architecture

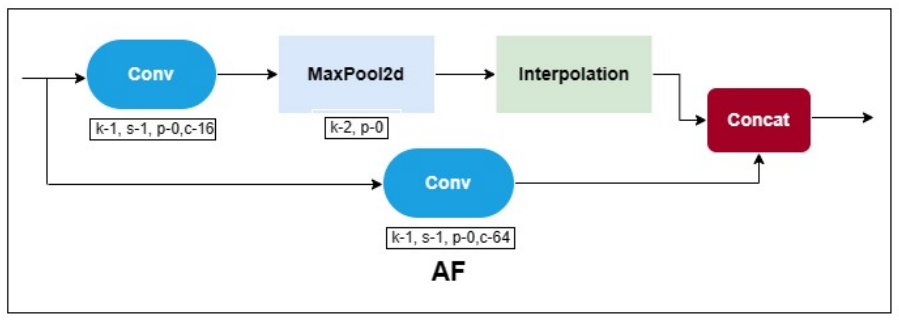

Fig. 3. Implementation of AF Module in the enhanced YOLOv5

## III. Results and Discussion

The suggested pipeline was trained and tested on a dataset consisting of 578 images obtained from patients, cadavers, and phantoms. This dataset was divided into 500 training images, 58 validation images, and 20 test images., with a specific emphasis on enhancing the YOLOv5 algorithm for vertebrae segmentation and testing the GUI for image-based screw placement and planning. The inference time for the vertebrae segmentation with Yolov5 was averaged at 2.8 ms on an high end i-7 laptop with 8 GB graphical support. The screws are initialized based on the output of the segmentation algorithm. The surgeon subsequently plans the remaining steps for the desired surgery. Due to AP-LP correspondence and synchronization, the surgeon only needs to plan on either of the AP or LP X-ray images, and the plan will automatically be updated on the other image. The C-Arm manufacturers used for this study is detailed in Table (I).

### A. Vertebrae Segmentation

Fig (6) shows the outcomes of spine vertebrae segmentation using various models. The models that were compared included YOLOv5, AF module with YOLOv5, and Spine-Vision model (YOLOv5+AF + CA). The output of each model was superimposed on the original image, and a confidence score was provided for each segmented vertebra. In Fig (6)a, the confidence score for the YOLOv5 model was 0.93, signifying a high level of accuracy in identifying and segmenting the vertebrae. However, there were instances where the model failed to accurately segment the vertebrae. Subsequently, In

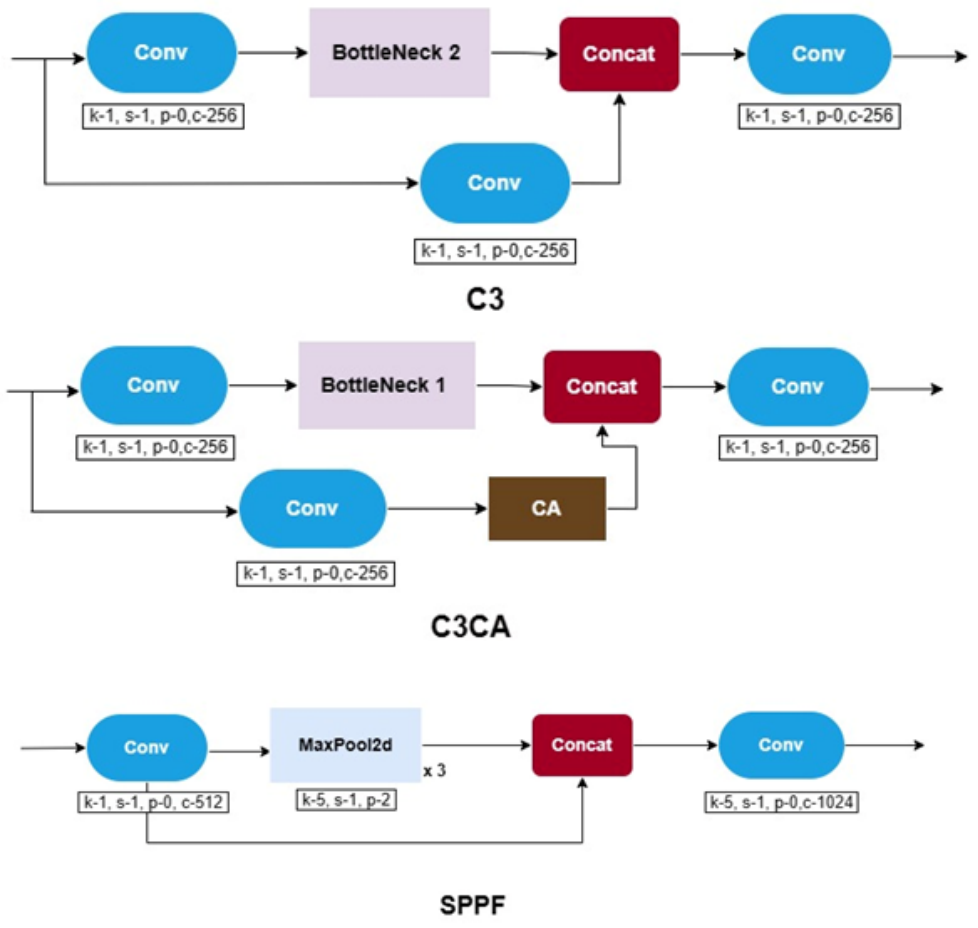

Fig. 4. Execution of C3CA & SPPF Module in the Enhanced YOLOv5

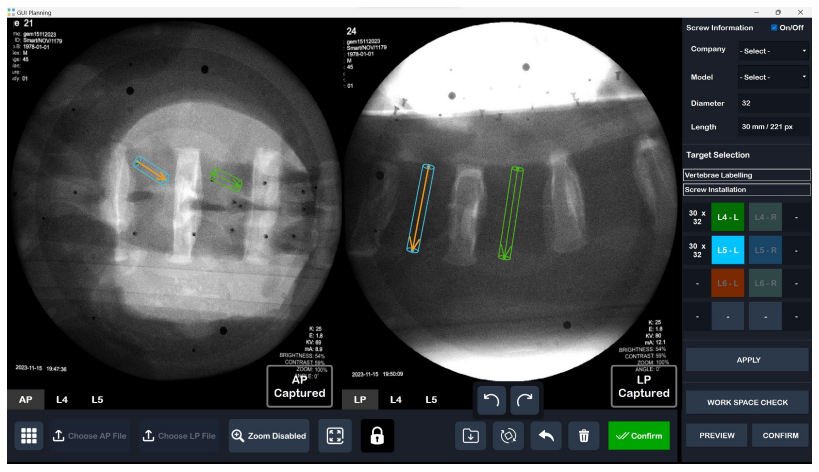

Fig. 5. Optimal positioning and strategic arrangement of screws on AP and LP images

TABLE I
DIFFERENT C-ARM'S USED FOR TESTING OF THE PROPOSED GUI FRAMEWORK

| SNO | C-Arm Model | C-Arm Manufacturer |
|---|---|---|
| 1 | GE-OEC-Fluorostar | GE-OEC |
| 2 | Ziehm 8000 | Ziehm |
| 3 | Ciosfit | Siemens |

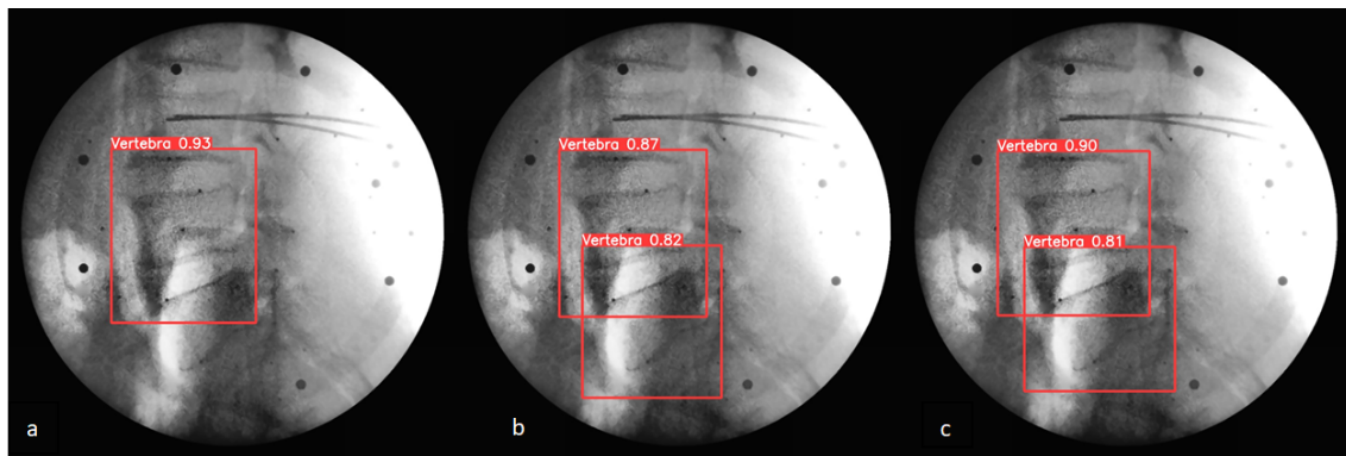

Fig. 6. (a) YOLOv5 (b) YOLOv5AF module (c) Spine-Vision

TABLE II
PERFORMANCE COMPARISON OF DIFFERENT DEEP LEARNING METHODS FOR VERTEBRAE SEGMENTATION WITH IN HOUSE TRAINING DATA OF 578 IMAGES

| Model Name | Precision | Recall | mAP @ 0.5 | mAP @ 0.5:0.95 |
|---|---|---|---|---|
| YOLOv3 | 0.9311 | 0.9732 | 98.17 | 80.62 |
| YOLOv7 | 0.9482 | 0.9821 | 98.18 | 80.65 |
| YOLOv8 | 0.9594 | 0.9821 | 98.56 | 83.3 |
| YOLOR | 0.9474 | 0.9554 | 98.23 | 83.05 |
| YOLOv6 | NA | NA | 96.59 | 0.785 |
| EfficientDet | NA | NA | 80.85 | 59.78 |
| YOLOv5 | 0.93128 | 0.98214 | 98.66 | 83.8 |
| Spine-Vision | 0.9417 | 0.96429 | 98.215 | 80.9 |

Fig (6)b, the AF module added to YOLOv5 yielded confidence scores of 0.81 and 0.90, slightly lower than those of the YOLOv5 model. Nevertheless, the addition of AF to YOLOv5 resulted in the segmentation of another vertebra that was not identified by YOLOv5 alone. In Fig (6)c, the Spine-Vision model exhibited higher confidence scores than the preceding models such as YOLOV8 and YOLOVR which led to selection of the proposed YOLOV5 for our application. Spine-Vision was compared with other versions of YOLO to quantify its performance and ability to detect vertebrae by training on the similar vertebrate dataset with optimal parameters and the results are illustrated in Table (II). YOLOv5 achieved a superior mAP score in the confidence interval 0.5:0.95. While YOLOv5 showed the best class recall score, it fared relatively poorly concerning other models in model precision. Inference testing revealed that YOLOv5 and YOLOR show relatively better results for vertebrae detection. YOLOv5 was chosen as the base framework for Spine-Vision as it has lesser complexity in comparison with YOLOR. Spine-Vision achieved better precision than YOLOv5 although the mAP scores at 0.5:0:95 confidence score is lower than YOLOv5's 83.8 mAP. Although the mAP scores are slightly lower, Spine-Vision can perform better than YOLOv5 for vertebrae detection, especially for noisy, low-resolution X-ray images.

## *B. GUI Based Pedicle Screw Planning*

The graphical user interface developed for X-Ray image-based surgical screw planning. The orientation of selected AP & LP images may vary depending on the positioning of the device during X-ray imaging of the spine from top and side views. However, for geometric accuracy, the spine/vertebrae should be aligned vertically in the images. The user has to verify the orientation, following which both the images are utilized as input for the Spine-Vision model to perform vertebrae segmentation. The model produces bounding boxes for the vertebrae in both images and stores the coordinates of the top-left and bottom-right corners of each bounding box in a separate text file within a folder. Subsequently, the user proceeds to choose the vertebra and accurately annotates the vertebrae in both images by clicking on the corresponding plots. If a matching bounding box is identified, it is assigned a label indicating the vertebra name and saved in a map. If no matching bounding box is identified after iterating through all the bounding boxes, a pop-up notification is sent to the user, indicating that no comparable bounding box is found. The bounding box with a label is saved in a map, and a target marker image is shown at the clicked location, signifying successful labeling of the vertebra. The labeling process involves marking, for example, the L4 and L5 vertebrae, as seen in Fig (7). The visualization of a screw is shown as a cylindrical shape in two plots, as depicted in Fig (8). The issue of inconsistencies caused by imaging changes is resolved by synchronizing the position of the screw in the AP and LP planes. This is achieved by calculating the differences in Z coordinates between the target and entrance points in both planes. The Z values are adjusted by subtracting these variations, thus assuring precise alignment of the screw with the surgeon's specified locations. The surgeon can update the entry and target locations, enabling the cylinder to change direction in response to the user's movements on the plot, thereby maintaining a continuous movement.

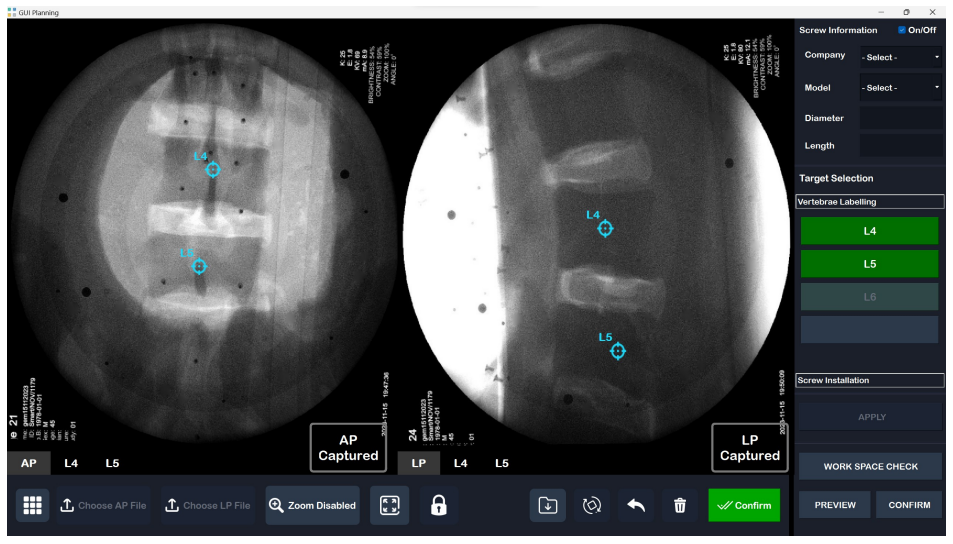

Fig. 7. Lumbar (L4 & L5) vertebrae were labelled

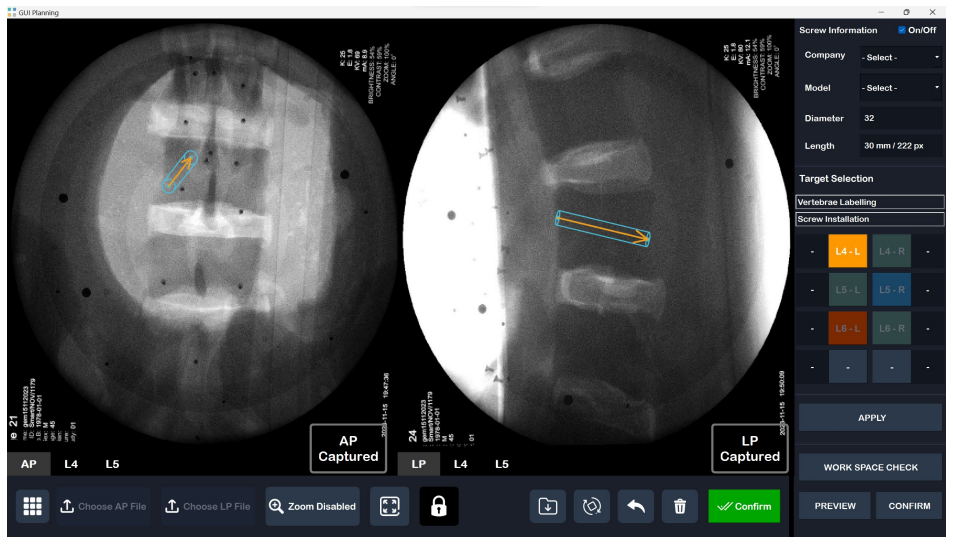

Fig. 8. L4 vertebra's Left screw is formed and visualized on both the plots

## IV. Conclusion

This work introduces an innovative GUI workflow for 2D X-ray image-based pedicle screw planning, aiming to enhance preoperative planning and intraoperative guidance. The Spine-Vision model, surpassing various YOLO iterations, including YOLOv5, in precision, recall, and mAP scores, excels in vertebrae segmentation. Particularly adept at detecting vertebrae in noisy, low-resolution X-ray images, Spine-Vision exhibits slightly lower mAP scores but compensates with high precision, ensuring accurate detection and segmentation. The GUI facilitates efficient management of spinal X-ray images, offering features for image orientation, vertebrae labeling verification, and seamless integration with the Spine-Vision model for segmentation. Planning involves vertebrae labeling, screw visualization, selection, and precise placement adjustment, enabling surgeons to achieve exact screw insertion in AP-LP images.